\documentclass[runningheads]{llncs}
\usepackage{graphicx}
\usepackage{amsmath,amssymb} 
\usepackage{color}
\usepackage[width=122mm,left=12mm,paperwidth=146mm,height=193mm,top=12mm,paperheight=217mm]{geometry}
\usepackage{multirow}
\usepackage{array}
\usepackage[hyphens]{url}
\usepackage[margin=0.5cm, skip=5pt]{caption}
\usepackage{makecell}
\begin{document}
\newcommand{\etal}{\textit{et al}.}
\newcommand{\ie}{\textit{i}.\textit{e}.}
\newcommand{\eg}{\textit{e}.\textit{g}.}
\newcolumntype{C}{>{\centering\arraybackslash}p{3.5em}}
\makeatletter
\newcommand{\printfnsymbol}[1]{%
  \textsuperscript{\@fnsymbol{#1}}%
}
\pagestyle{headings}
\mainmatter

\title{Egocentric 6-DoF Tracking of Small Handheld Objects} 

\titlerunning{}

\authorrunning{R. Pandey, P. Pidlypenskyi, S. Yang, C. Kaeser-Chen}

\author{Rohit Pandey\thanks{Equal contribution}, Pavel Pidlypenskyi\printfnsymbol{1}, Shuoran Yang, Christine Kaeser-Chen}


\institute{Google Inc.\\
	\email{ \{rohitpandey,podlipensky,shuorany,christinech\}@google.com}
}

\maketitle

\begin{abstract}

Virtual and augmented reality technologies have seen significant growth in the
past few years. A key component of such systems is the ability to track the
pose of head mounted displays and controllers in 3D space. We tackle the
problem of efficient 6-DoF tracking of a handheld controller from egocentric
camera perspectives. We collected the HMD Controller dataset which consist of
over 540,000 stereo image pairs labelled with the full 6-DoF pose of the
  handheld controller. Our proposed SSD-AF-Stereo3D model achieves a mean
  average error of 33.5 millimeters in 3D keypoint prediction and is used in
  conjunction with an IMU sensor on the controller to enable 6-DoF tracking. We
  also present results on approaches for model based full 6-DoF tracking.  All
  our models operate under the strict constraints of real time mobile CPU
  inference.

\end{abstract}

\section{Introduction}

In the past few years, virtual reality (VR) systems have seen an increased
demand.
These devices are typically in the form of a head mounted display (HMD)for
rendering the virtual scene, and single or dual handheld controllers for
interaction. The HMD and controllers need to be tracked in position and
orientation to create an immersive experience. The tracking can either be 3
degrees of freedom (DoF) including only orientation (roll, pitch and yaw) or
6-DoF which includes position in 3D space as well.

More realistic experiences can be created with 6-DoF tracking but it often
requires additional hardware. VR headsets like the HTC Vive use external IR
cameras and markers for tracking, restricting the system to be
only operational in limited space. Newer mobile 6-DoF headsets can achieve
similar results with inside out tracking. Such headsets have one or more outward
facing cameras attached to the headset. By applying localization algorithms such
as SLAM on camera images, we can compute the headset's 6-DoF position with
respect to the environment.

Meanwhile, tracking handheld controllers in 6-DoF for mobile HMD remains
a difficult problem. Controllers tend to move faster than headsets, have a much
larger movement range, and may be occuluded by users' own bodies. Existing
solutions rely on either additional sensing hardware, e.g. electromagnetic
sensors in Sixense systems, or additional visual markers as in Sony PS VR
systems. The former solution can be costly, and the latter suffers
from reliability issues when markers are occluded.

In this work, we explore image-based markerless 6-DoF tracking of handheld
controllers. Our key observation is that users' hands and arms
provide excellent context for where the controller is in the image, and
are robust cues even when the controller itself might be occluded. To simplify
the system, we use the same cameras for headset 6-DoF pose tracking on
mobile HMDs as our input. In our experiments, they are a pair of stereo
monochrome fisheye cameras. We do not require additional markers or hardware
beyond a standard IMU based controller. We believe this can enable extremely small
and cheap clicker-like controllers, and eventually lead into purely hand based interaction.

\subsection{Contributions}
Our main contributions in this work are:
\begin{enumerate}
  \item An approach to automatically label 6-DoF poses of handheld objects in camera
    space.
  \item The HMD Controller dataset\footnote{https://sites.google.com/view/hmd-controller-dataset}, 
    the largest-to-date markerless object 6-DoF
    dataset.  This dataset contains 547,446 stereo image pairs with the
    6-DoF pose of a handheld controller. We provide timestamped 6-DoF pose of a
    handheld controller for each image.  The dataset contains images for 20
    different users performing 13 different movement patterns under varying
    lighting conditions. Our dataset will be publicly available prior to the
    conference.
  \item Neural network models to enable 3-DoF and 6-DoF tracking of handheld
    objects with mobile CPU compute constraints.
\end{enumerate}
\section{Related Work}

There are a few existing datasets of handheld objects
for the task of object recognition. The Small Hand-held Object Recognition Test
(SHORT) \cite{Rivera-Rubio2014} dataset has images taken from hand-held or
wearable cameras. The dataset collected in \cite{liu2014multiple} uses RGBD data
instead. The Text-IVu dataset \cite{beck2014text} contains handheld objects
with text on them for text recognition. None of these datasets contain pose
information of handheld objects.

On the other hand, researchers have also collected datasets specifically for
object 6-DoF pose estimation. Datasets like the ones presented in
\cite{hinterstoisser2012model} and \cite{tejani2014latent} provide full 6D
object pose as well as the 3D models for most object categories, but do not deal
with handheld objects. It is worth noting that having 3D models of the objects
can improve pose estimation accuracy, but is infeasible in our
case where the hand shape and manner of holding the object vary across users.

Our work is closely related to hand pose estimation from an egocentric
perspective. The EgoHands dataset \cite{Bambach_2015_ICCV} consists of videos
taken from a Google Glass of social interactions between people. It contains
pixel level segmentation of the hands, but no information on handheld objects.
The SynthHands dataset \cite{OccludedHands_ICCV2017} consists of real captured
hand
motion retargeted to a virtual hand with natural backgrounds and interactions
with different objects. The BigHand2.2M benchmark dataset
\cite{yuan2017bighand2} is a large dataset which uses 6D magnetic sensors and
inverse kinematics to automatically obtain 21 joints hand pose annotations on
depth maps. The First-Person
Hand Action Benchmark dataset \cite{garcia2017first}
provides RGB-D video sequences of the locations of 21 joints on the hand as well
as the 6-DoF pose of the object the hand is interacting with. The joint
locations are captured using visible magnetic sensors on the hand.

We base our models on the SSD \cite{liu2016ssd} architecture due to its
computational efficiency and superior performance compared to other single shot
object detection approaches like YOLO \cite{redmon2016you}. Some of the key
factors for the improved accuracy of SSD come from using separate filters for
different aspect ratios. They are applied to feature maps at different feature
extractor layers to perform detection at multiple
scales. The
computational efficiency comes from the fact that it is a single stage approach
that does detection and recognition in one go instead of two stage approaches
like Faster RCNN \cite{girshick2015fast} and Mask RCNN \cite{he2017mask} that do
detection in the first stage followed by recognition in the second. 

A few recent work have extended object detection approaches to 3D and 6D in a
manner similar to ours.  Mousavian \etal \cite{mousavian20173d} regress box
orientation and dimensions for 3D bounding
box prediction on the KITTI \cite{geiger2012we} and Pascal 3D+
\cite{xiang2014beyond} datasets.  Kehl \etal \cite{kehl2017ssd} and Poirson
\etal \cite{poirson2016fast} have formulated object pose estimation as
a classification problem of discrete pose candidates, and tackled the problem using
variants of SSD.  Other approaches like \cite{mahendran20173d} treat pose
estimation as a regression problem, and 
used a combination of a CNN based feature network and an object specific pose
network to regress 3D pose directly.

\section{HMD Controller Dataset}
\label{sec:dataset}

The HMD Controller dataset consists of over 540,000 stereo monochrome fisheye
image pairs of 20 participants performing 13 different movement patterns with
the controller in their right hand. We collect the 6-DoF pose (position and
orientation in 3D space) of the handheld controller. For each image pair sample,
we provide:

\begin{itemize}
    \item Timestamp synchrnoized 6-DoF pose of the tip of the controller in left
      camera's coordinate frame;
    \item Timestamp synchronized 6-DoF pose of the cameras with respect to a
      static environment;
    \item Intrinsics and extrinsics of the camera pair.
\end{itemize}

\subsection{Device Setup and Calibration}

To collect precise groundtruth for 6-DoF pose of the controller we use the Vicon
motion capture system \cite{vicon_online}, which can track retroreflective markers
with static infrared cameras set up in the capture space. We attach
constellation of Vicon markers to both the headset and the controller. The
markers on the headset are outside the field of view of cameras.  The markers on
the controller are placed in a way that they would be occluded by human hand
most of the time, and not visible in camera images. We asked users to perform
predefined set of motions. As there was still risk to have certain poses where
markers were visible, we had three versions of controllers with Vicon markers
attached to different places (Fig \ref{fig:paprikas}). Users were asked to
repeat the same motion with each version of the controller.

\begin{center}
\begin{minipage}[t]{.495\linewidth}
\vspace{0pt}
\raggedright
\includegraphics[width=\textwidth]{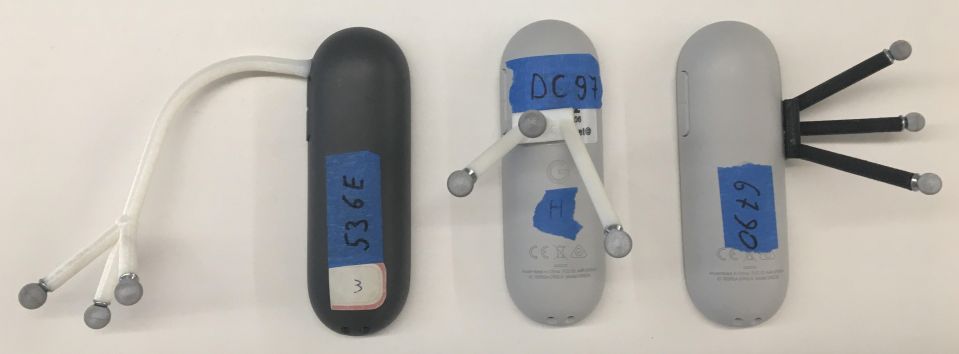}
\captionof{figure}{Different versions of Vicon attachments to controller.}
\label{fig:paprikas}
\end{minipage}%
\begin{minipage}[t]{.48\linewidth}
\vspace{0pt}
\centering
\begin{tabular}{|c|c|c|}
\hline
                       & \thead{Orientation \\ (degree)} & \thead{Position \\ (mm)} \\ \hline
  \thead{Headset}    & 0.349              & 6.693       \\ \hline
  \thead{Controller} & 0.032           & 0.658       \\ \hline
\end{tabular}
\captionof{table}{RMSE in hand-eye calibrations.}
\label{tbl:calibration-errors}
\end{minipage}
\end{center}

Vicon system provides 6-DoF pose tracking at 500 Hz. The pose of
the marker constellation on the back of the controller $CB$ with respect to the Vicon
room origin $V$ (initialized in an one-off room calibration stage) is provided at every
update. We denote this as $T_{V}^{CB}$. 

As we have different marker constellations, we need to compute the
pose of one canonical keypoint on the controller to be able to merge data
captured in different sessions. We choose the tip
of the controller $CT$ as the canonical keypoint, as it best reflects users'
intention for spatial interaction in VR. We futher define the local coordinate
space centered at controller tip to be axis-aligned with the physical
controller. We denote the transformation between
this canonical coordinate space and the camera space as $T_{Cam}^{CT}$.

The computation of $T_{Cam}^{CT}$ depends on the tracking of headset in vicon space. We track the headset by introducing an additional Vicon trackable
constellation $H$ which is mounted rigidly on the headset.  At every frame we
receive updates on the pose of $H$ in Vicon tracking space. We
denote this as $T_{V}^{H}$. The headset-mounted
constellation has a rigid transformation from the camera. This can be computed as a hand-eye calibration problem \cite{chou1991finding} in robotics.

We compute the rigid transformation of $T_{Cam}^{H}$ with
offline hand-eye calibration procedures. We also compute $T_{CB}^{CT}$ offline,
with temporarily mounting a tracking marker at the controller tip and record the
Vicon poses of $CB$ and $CT$:
\begin{equation}
  T_{CB}^{CT} = T_{V}^{{CB}^{-1}} \cdot T_{V}^{CT}.
\end{equation}

After $T_{CB}^{CT}$ is computed for each controller configuration, we remove the
marker on the tip of the controller for user data collection in order not to
introduce visible markers to images. 

Another important calibration step is time alignment between Vicon clock and
headset camera clock. The alignment is done based on angular velocity calculated
based on trajectory provided by Vicon and cameras handeye calibration
\cite{chou1991finding}. This allows us to find camera frame and corresponding
6-DoF pose of the controller.

During data collection, we can compute synchronized 6-DoF pose of the tip of the
controller in left camera space as follows:
\begin{equation}
T_{Cam}^{CT} = T_H^{{Cam}^{-1}} \cdot T_V^{H^{-1}}\cdot T_V^{CB} \cdot T_{CB}^{CT}. 
\end{equation}

Each of the calibration steps described above introduced some error, we estimate
root mean squared errors (RMSE) for hand-eye calibration in Table
\ref{tbl:calibration-errors}. This represents the level of noise in groundtruth
labels of our dataset.  

\subsection{Dataset Cleaning}

We investigated two potential issues with samples in our dataset: frames
with missing or incorrect 6-DoF poses, and frames with visible tracking markers.

To remove frames with missing or incorrect 6-DoF poses, we filter our dataset
with the following criteria:

\begin{enumerate}
\item Controller position is restricted to be within 1 meter away from
  the camera. All our dataset participants have arm length less than 1 meter.
\item We can detect missing Vicon tracking frames with the Vicon system. We
  discard image frames with no corresponding Vicon poses. We also note that it
  takes approximately 0.6s for Vicon to fully re-initialize. Poses produced
  during the reinitialization stage tends to be erroneous. Therefore we 
  discard 20 subsequent camera frames after tracking is lost as well.
\item Incorrect 6-DoF labels due to Vicon tracking errors are more difficult to
  filter automatically. Figure \ref{fig:vicon-failures} provides a few examples
    of imprecise pose labels. We use an active learning scheme for filtering
    such frames, where we apply our trained models on the dataset to detect
    potential incorrectly labeled frames. In our experiments, we detect frames
    with pose prediction error larger than 3cm. We then manually scan the
  set of frames and remove invalid labels. 
\end{enumerate}

We provide timestamps for each image in our cleaned dataset, so that one can
track discontinuity in frames.

Another potential issue with our dataset is accidental exposure of Vicon
tracking markers in images. Since users are encouraged to move freely when
completing a motion pattern, it is always possible that tracking markers are
visible to head-mounted cameras.

We use the integrated gradient method \cite{sundararajan2017axiomatic} to
analyze effects of visible markers in the training dataset. Interestingly, we
observe that our train models do not pick up markers as visual cues. In Figure
\ref{fig:integrated-gradient}, we show an sample input image with clearly
visible markers. Pixels that contributed most to model prediction on this image
does not include pixels of the marker. We believe this is due to the small number of
frames with visible markers and the small size of markers in the grayscale image
- each marker is approximately 2.5 pixels wide on average in training images.
Therefore, our final dataset does not explicitly filter out frames with visible markers.

\begin{center}
  \centering
  \begin{minipage}[b]{0.495\textwidth}
  \vspace{0pt}
  \begin{minipage}[b]{0.48\textwidth}
    \includegraphics[width=\textwidth]{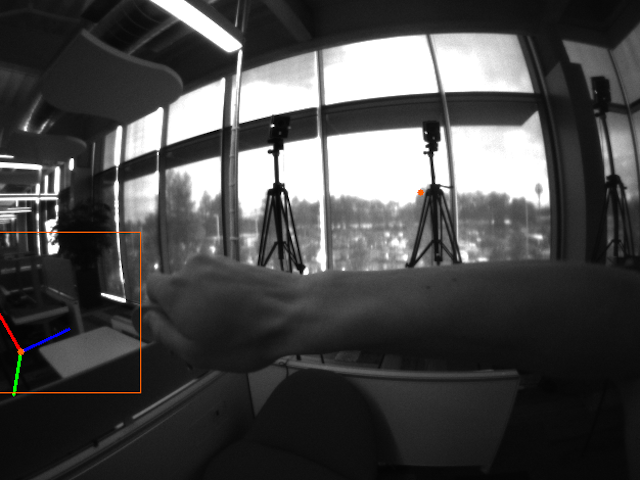}
  \end{minipage}
  \begin{minipage}[b]{0.48\textwidth}
    \includegraphics[width=\textwidth]{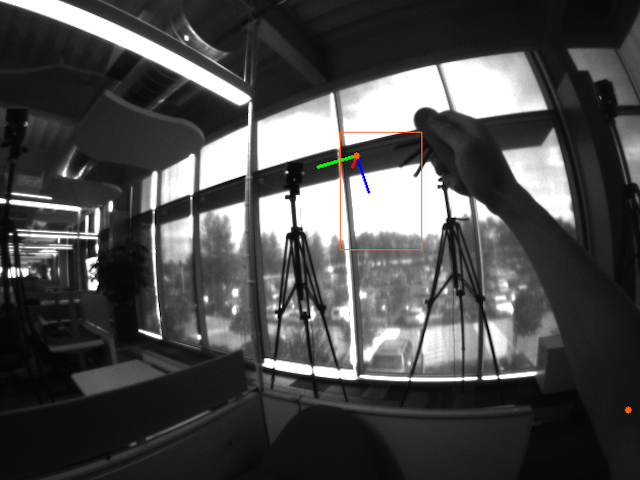}
  \end{minipage}
  \captionof{figure}{Examples of Vicon tracking failures and incorrect 6-DoF poses.}
  \label{fig:vicon-failures}
  \end{minipage}
  \begin{minipage}[b]{0.495\textwidth}
  \vspace{0pt}
  \begin{minipage}[b]{0.48\textwidth}
    \includegraphics[width=\textwidth]{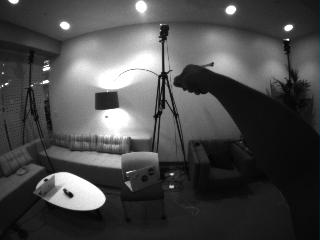}
  \end{minipage}
  \begin{minipage}[b]{0.48\textwidth}
    \includegraphics[width=\textwidth]{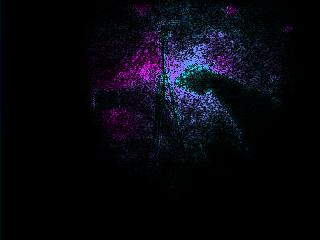}
  \end{minipage}
  \captionof{figure}{Input image with visible markers and pixels that contributed  most to prediction.}
  \label{fig:integrated-gradient}
  \end{minipage}
\end{center}

\subsection{Bounding Box and Label Assignment}

Many objection detection models such as SSD \cite{liu2016ssd} require 2D
bounding boxes and labels as input. Since the controller to be tracked is
largely occluded by hand, we instead compute the bounding box for the hand
holding the controller. We observe that we can approximate users' thumb position 
with the controller tip position $CT$. 

To compute the 3D bouding cube of users' hand in camera space, we have:
\begin{equation}
  P_{Cam}^{c_i} = T_{Cam}^{CT} \cdot P_{CT}^{c_i}  \quad \text{for } x =
  1,\dotsc,8
\end{equation}
where $c_i$ denotes the 8 corners of the bounding cube, and $P_{CT}^{c_i}$
denotes the location of the corners in local controller space $CT$. In our
experiments, we set $P_{CT}^{c_i}$ to be the permutation of $\{\{-0.03, 0.05\},
\{-0.05, 0.01\}, \{-0.01, 0.10\}\}$ (in meters). The bounding box size reflects
typical human hand size, and the location reflects the shape of the right hand
viewed from the controller tip's local space.

Finally we compute the 2D hand bounding box by projecting the 3D bounding cube
into the image space using camera intrinsics. We choose the smallest
axis-aligned 2D bounding box that contains all projected bounding cube corners.
All hand bounding boxes are automatically associated with the label for right
hand.  During model training, we add another label for background to all
unmatched anchors.

Note that with using the full hand as object bounding box, we have transferred
our problem to be hand detection and keypoint localization in image space. Hands and arms
provide excellent context for controller pose even when the controller is
not observable visually. Hands and arms also have more high-level features for neural
networks to pick up on. This is a key to our solution of markerless
controller tracking.

\subsection{Dataset Statistics}

After dataset cleaning, we obtained a final set with 547,446 frames. Figure
\ref{fig:samples} shows sample frames with visualized groundtruth pose annotations. Figure
\ref{fig:uv-heatmap}--\ref{fig:orientation-hist} shows pose distribution in our dataset in image space,
${xyz}$ space and orientation space. 

\begin{center}
  \centering
  \begin{minipage}[b]{0.235\textwidth}
    \includegraphics[width=\textwidth]{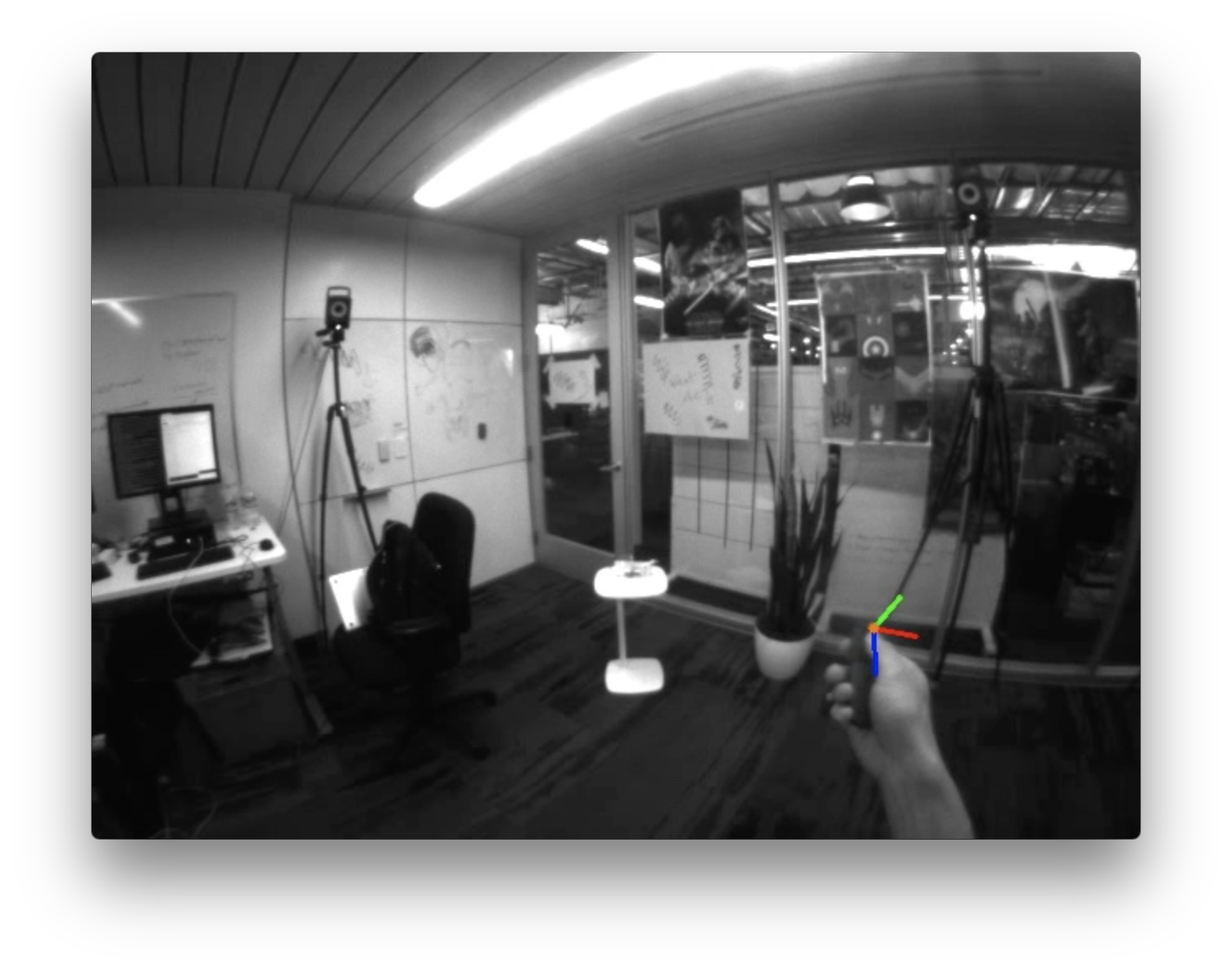}
  \end{minipage}
  \begin{minipage}[b]{0.235\textwidth}
    \includegraphics[width=\textwidth]{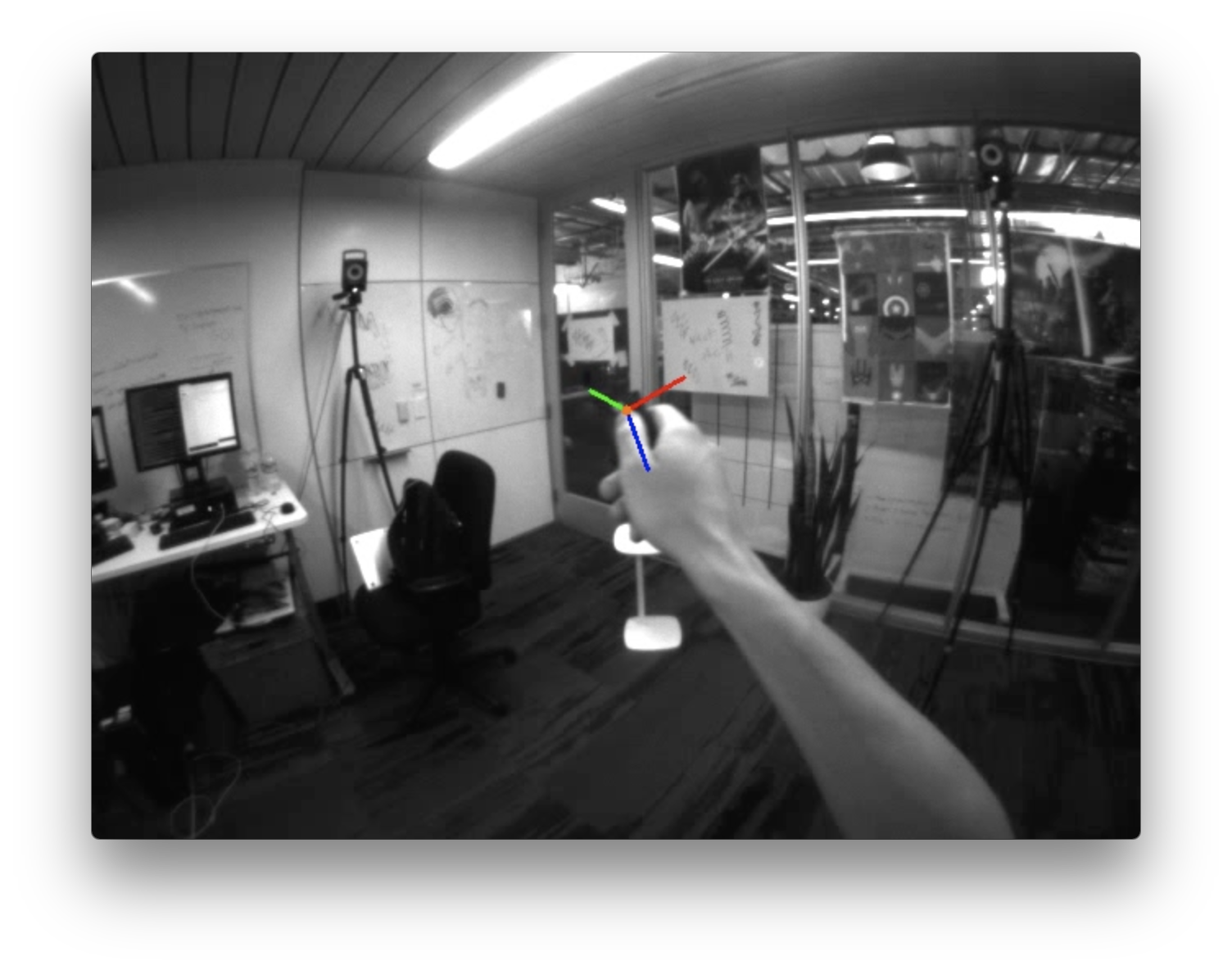}
  \end{minipage}
  \begin{minipage}[b]{0.235\textwidth}
    \includegraphics[width=\textwidth]{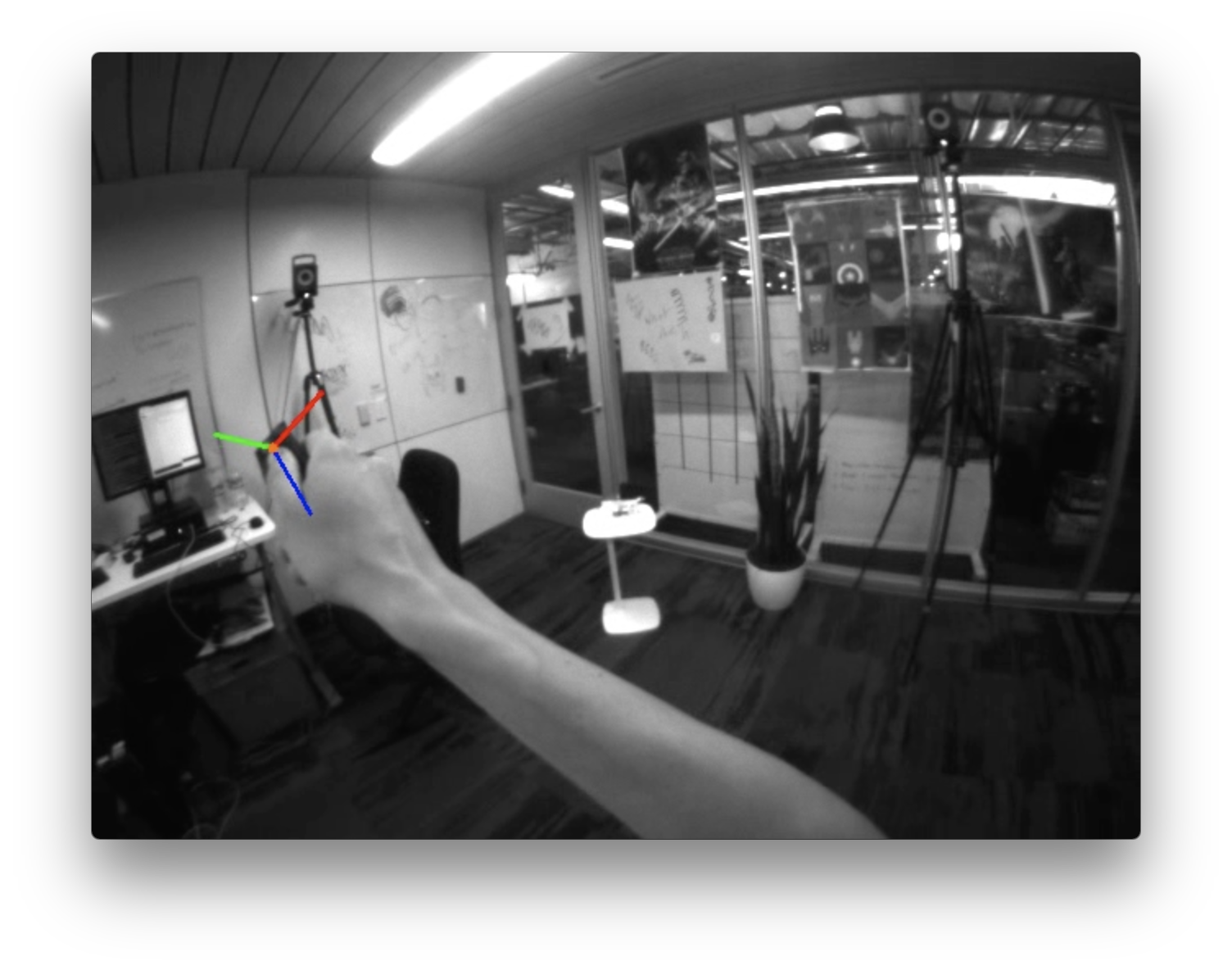}
  \end{minipage}
  \begin{minipage}[b]{0.235\textwidth}
    \includegraphics[width=\textwidth]{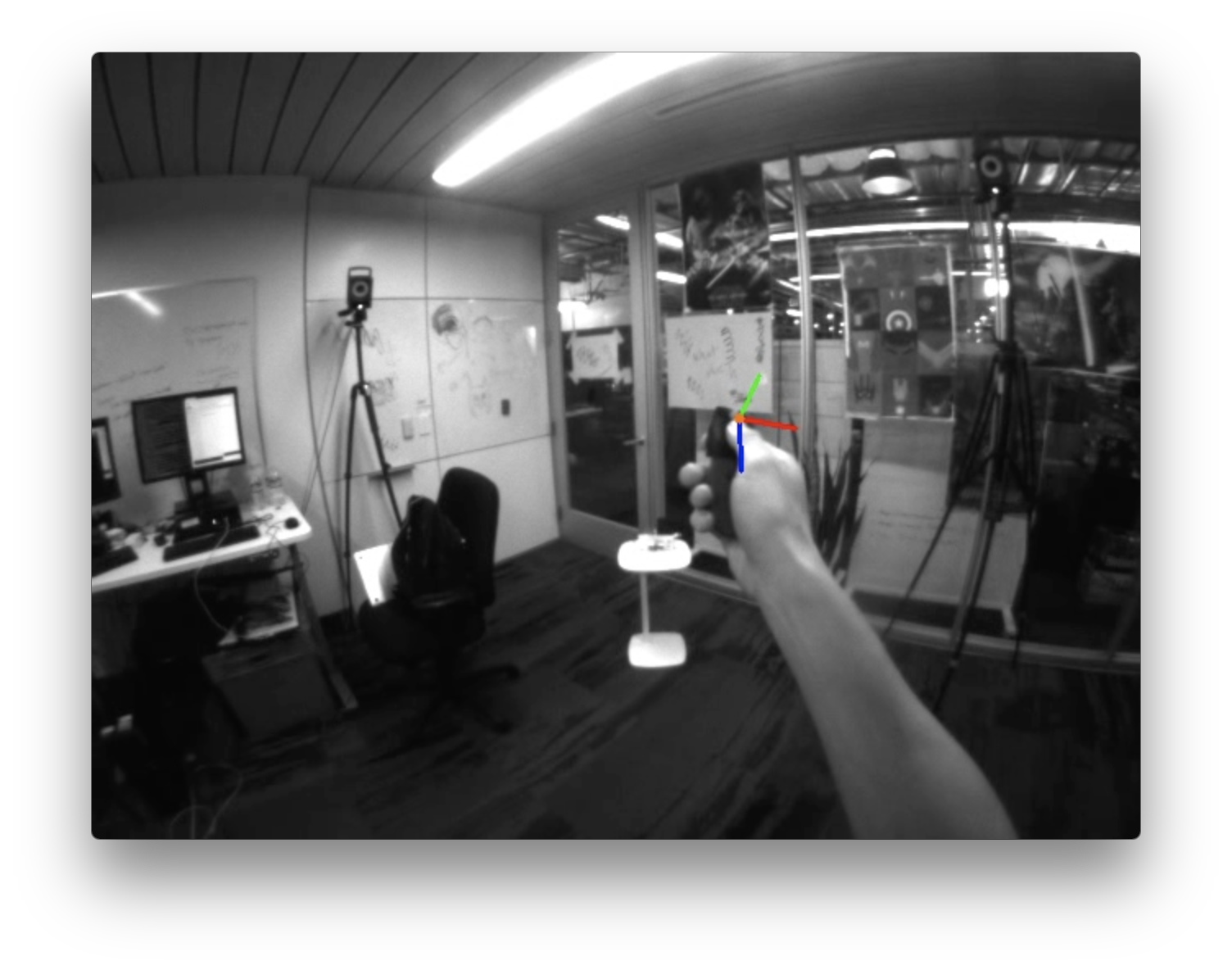}
  \end{minipage}
  \par 
  \begin{minipage}[b]{0.235\textwidth}
    \includegraphics[width=\textwidth]{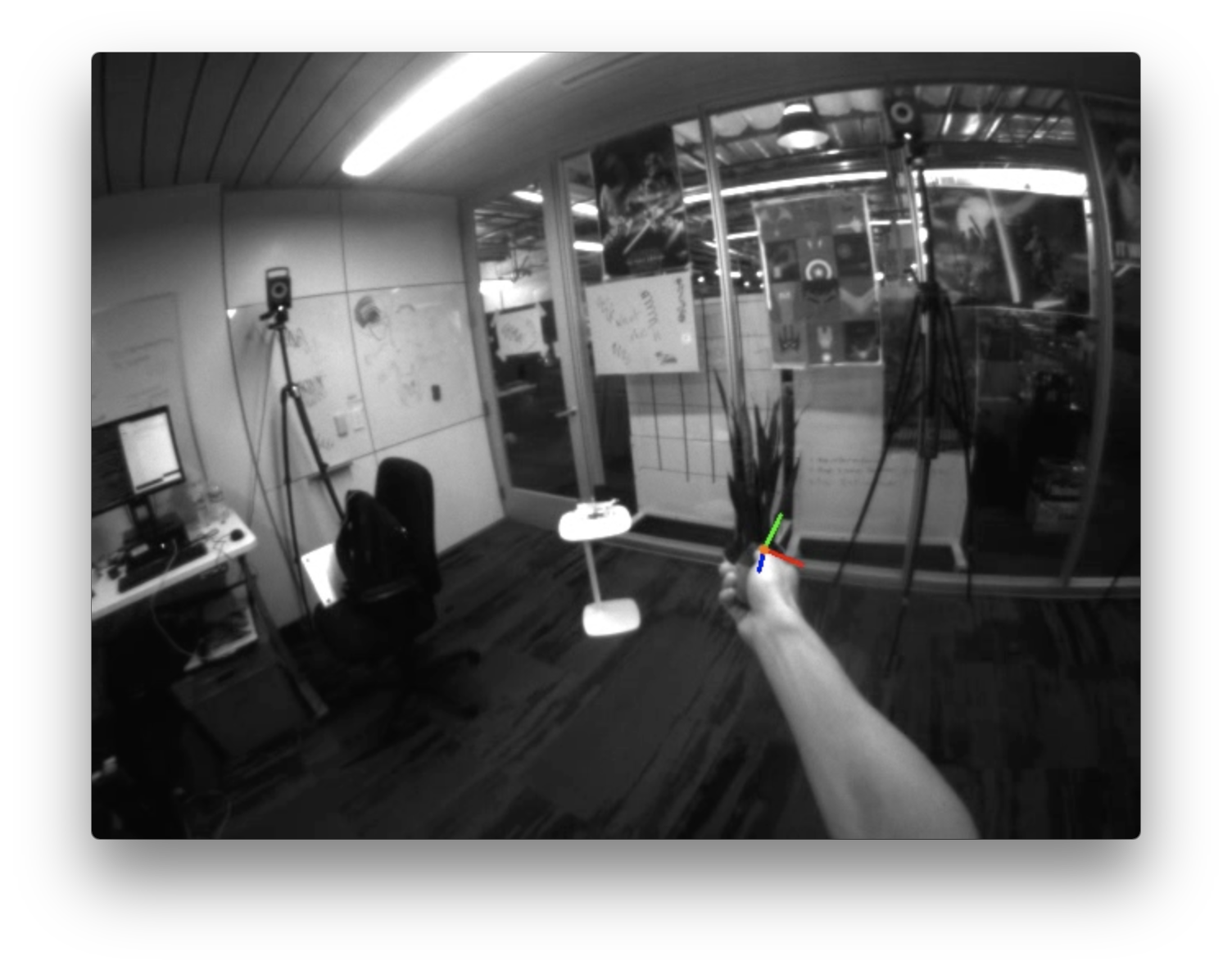}
  \end{minipage}
  \begin{minipage}[b]{0.235\textwidth}
    \includegraphics[width=\textwidth]{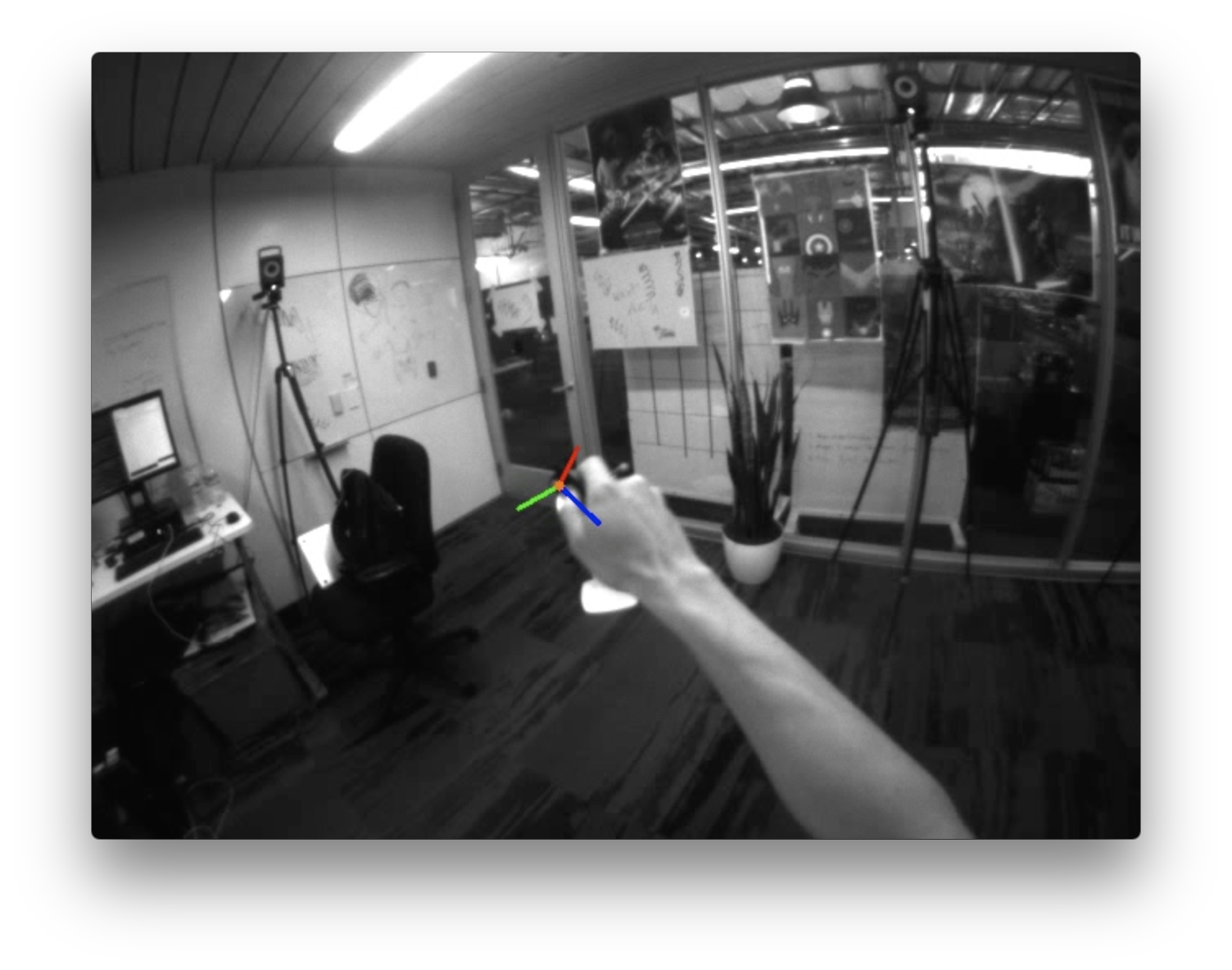}
  \end{minipage}
  \begin{minipage}[b]{0.235\textwidth}
    \includegraphics[width=\textwidth]{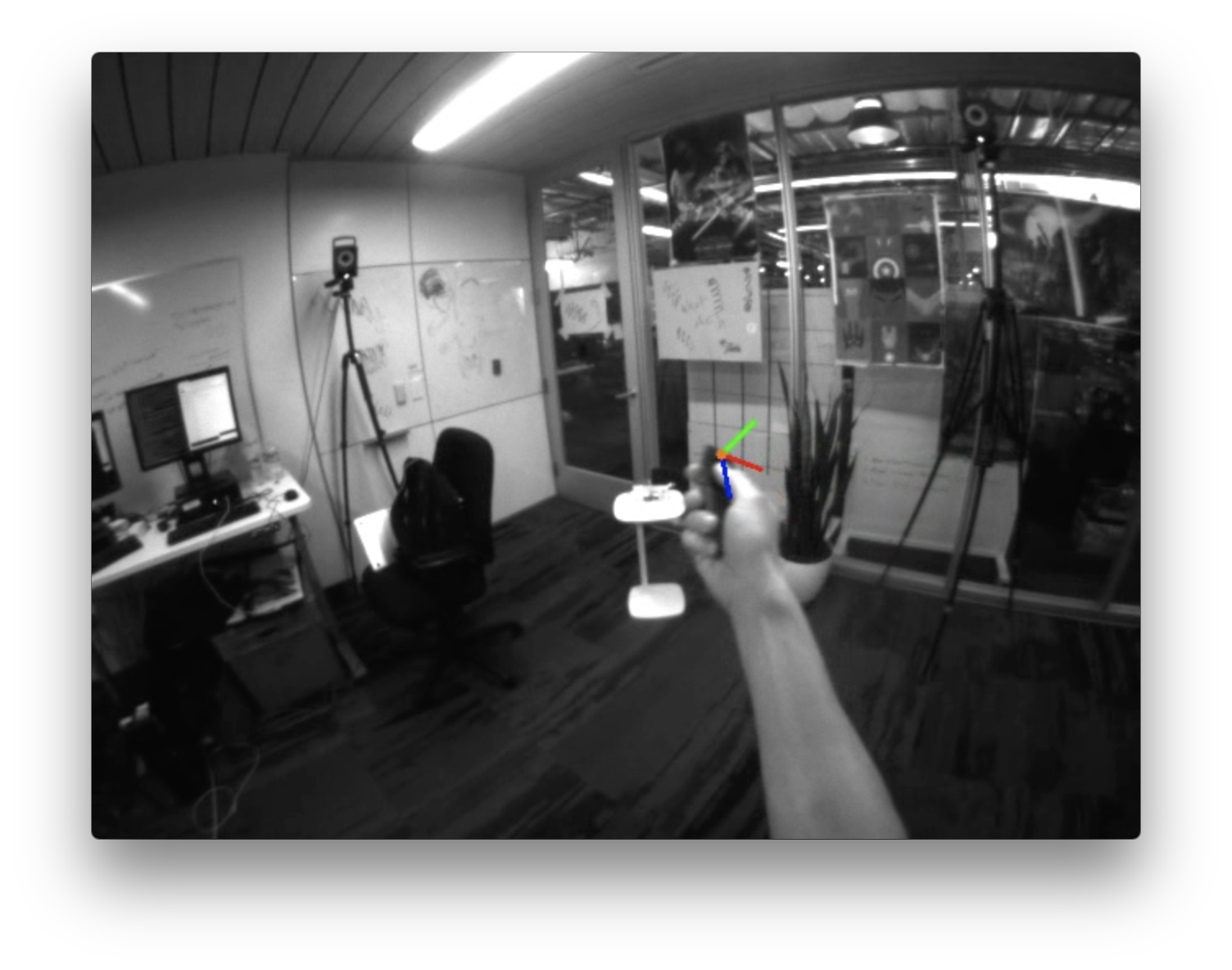}
  \end{minipage}
  \begin{minipage}[b]{0.235\textwidth}
    \includegraphics[width=\textwidth]{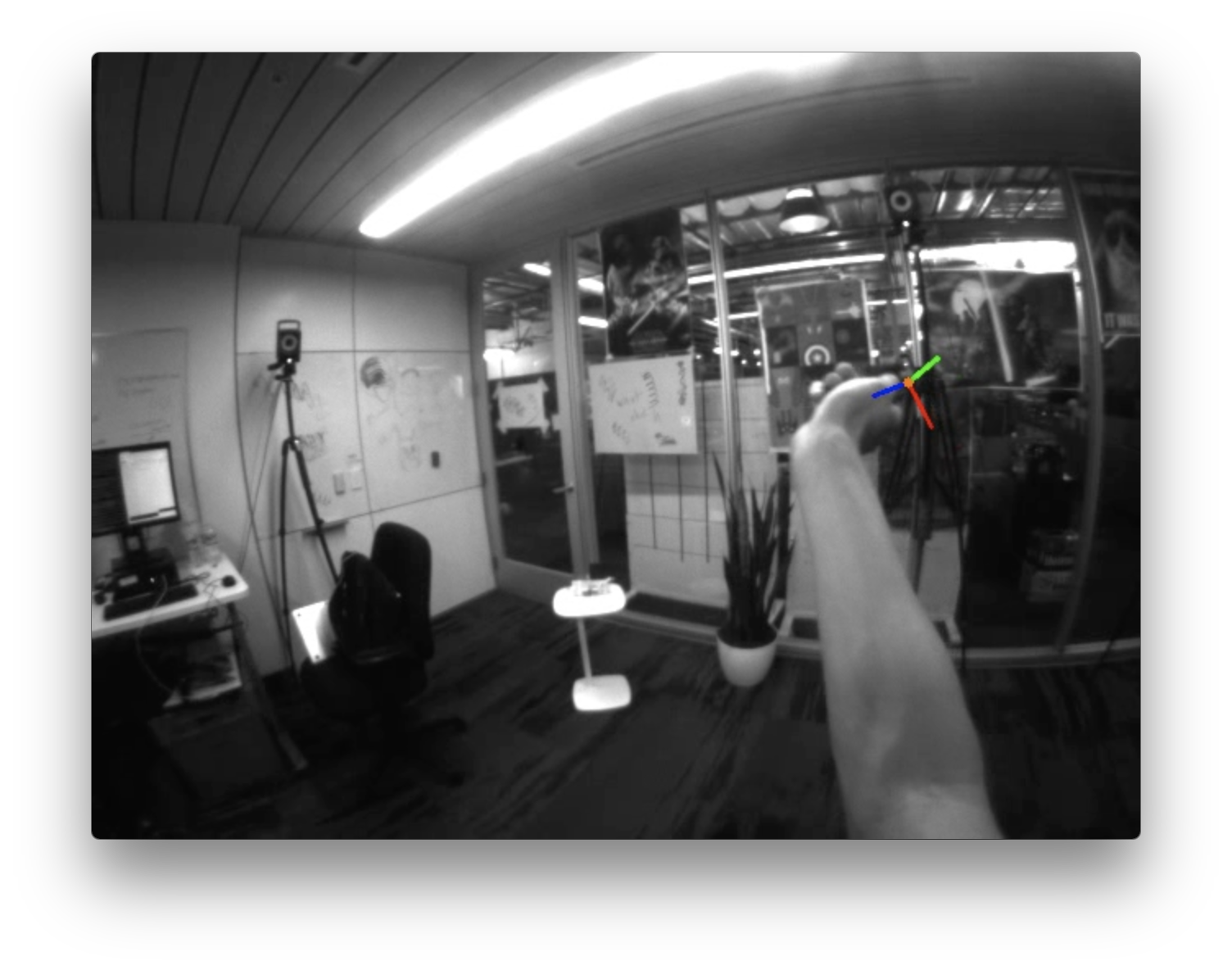}
  \end{minipage}
  \par
  \captionof{figure}{Sample images with visualized 6-DoF groundtruth annotation.}
  \label{fig:samples}
\end{center}

\begin{center}
  \begin{minipage}[b]{0.3\textwidth}
    \includegraphics[width=\textwidth]{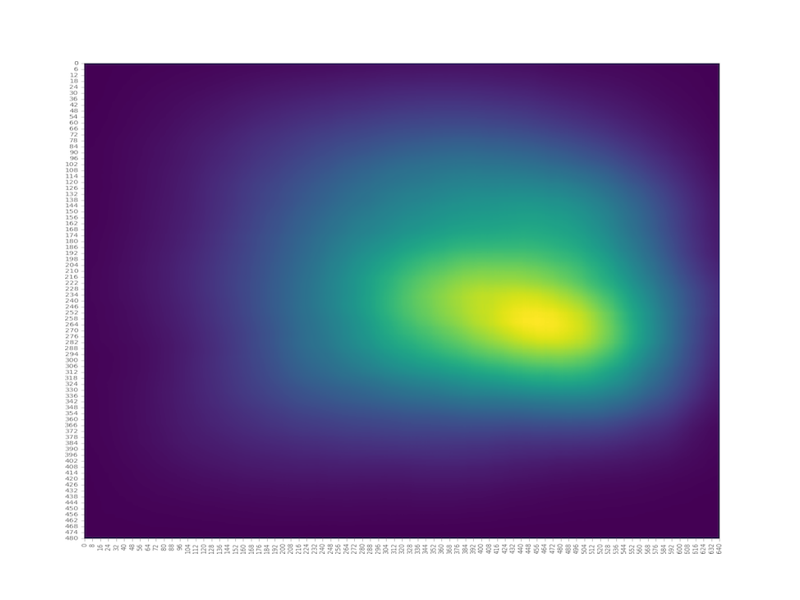}
  \end{minipage}
  \begin{minipage}[b]{0.3\textwidth}
    \includegraphics[width=\textwidth]{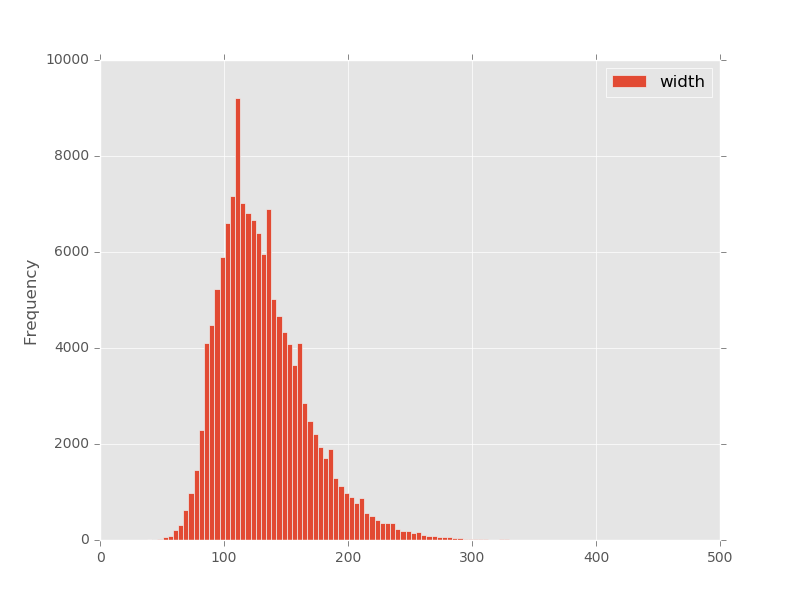}
  \end{minipage}
  \begin{minipage}[b]{0.3\textwidth}
    \includegraphics[width=\textwidth]{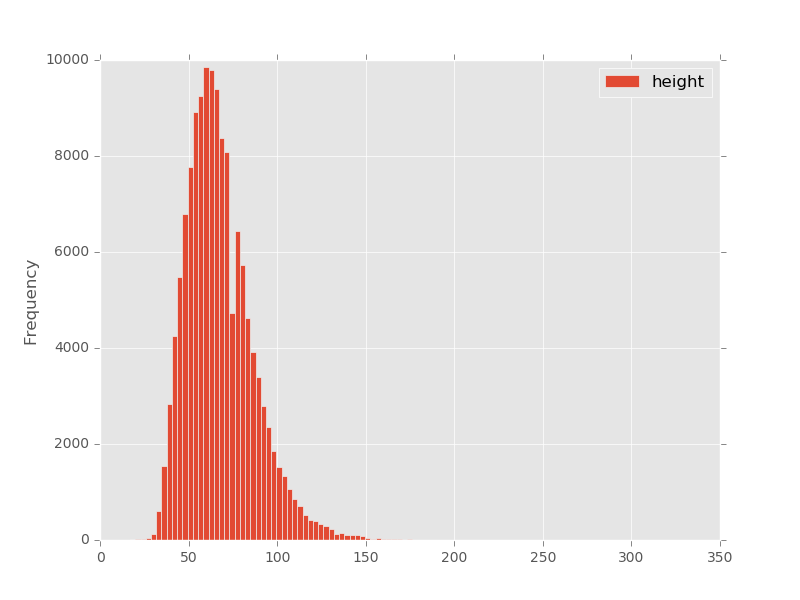}
  \end{minipage} \par
  \captionof{figure}{\textit{Left to right}: Heatmap of pixels occupied by user hand and
  controller; histogram of 2D bounding box width; histogram of 2D bounding box height.}
  \label{fig:uv-heatmap}
  \begin{minipage}[b]{0.3\textwidth}
    \includegraphics[width=\textwidth]{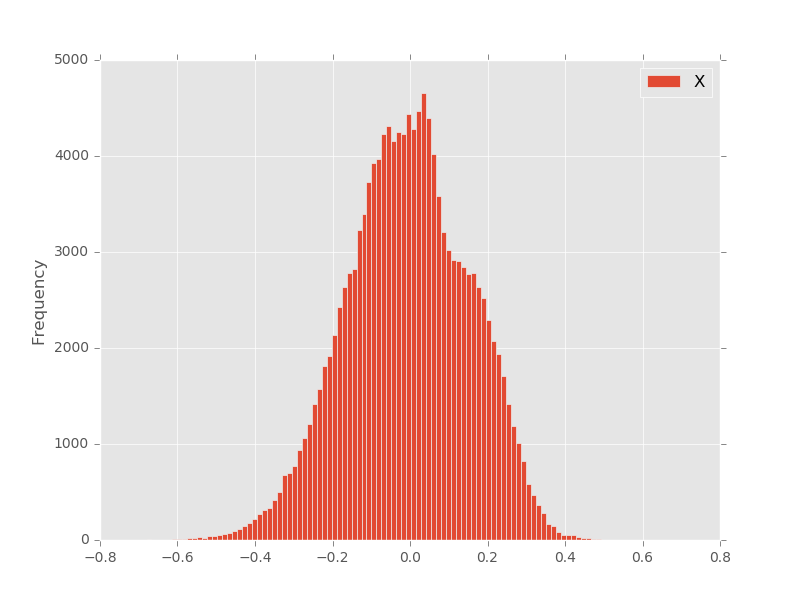}
  \end{minipage}
  \begin{minipage}[b]{0.3\textwidth}
    \includegraphics[width=\textwidth]{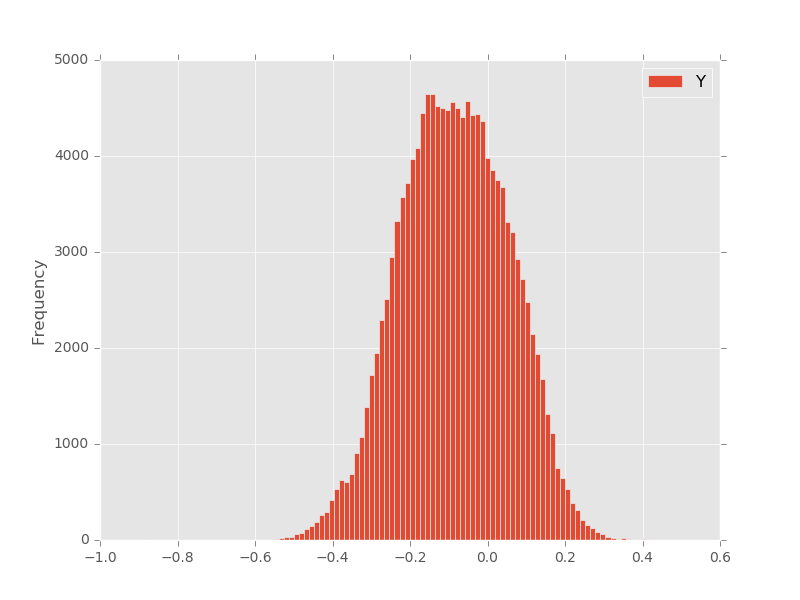}
  \end{minipage}
  \begin{minipage}[b]{0.3\textwidth}
    \includegraphics[width=\textwidth]{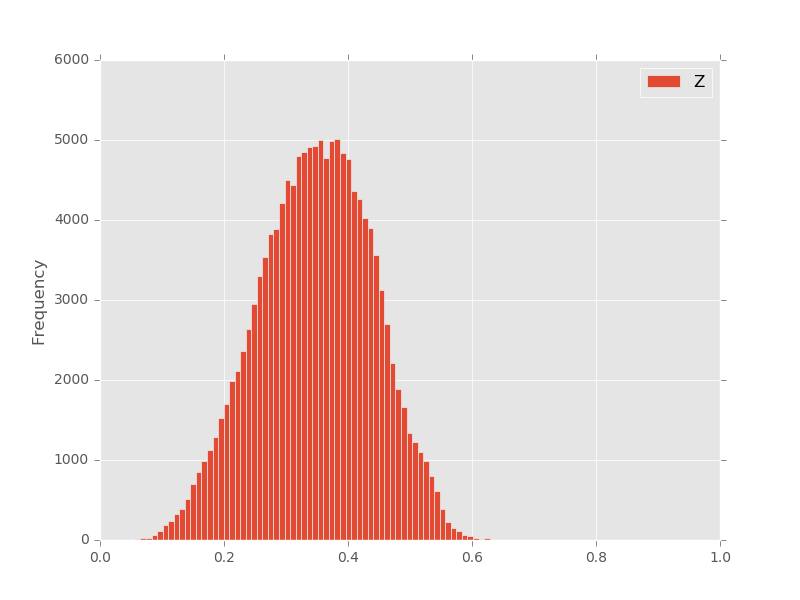}
  \end{minipage} \par
  \captionof{figure}{\textit{Left to right}: Histograms of coordinates in x, y
  and z directions in meters.}
  \label{fig:position-hist}
  \begin{minipage}[b]{0.3\textwidth}
    \includegraphics[width=\textwidth]{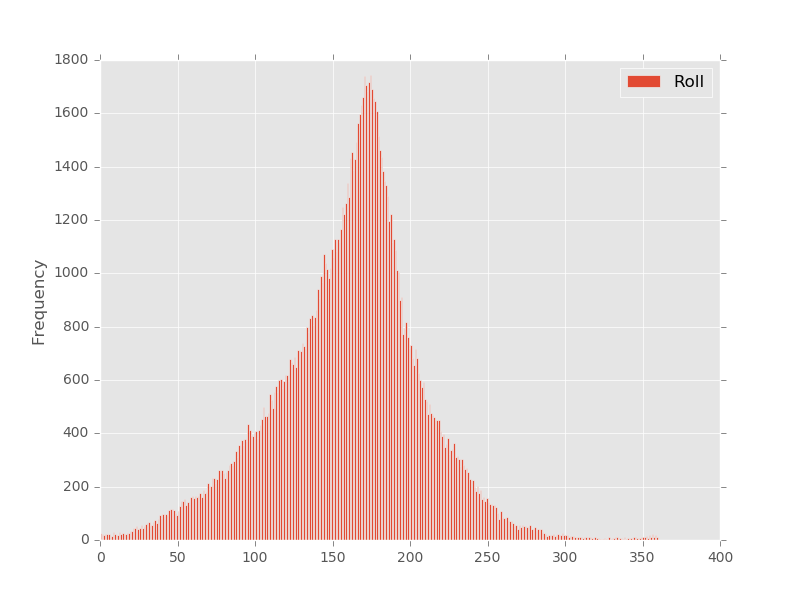}
  \end{minipage}
  \begin{minipage}[b]{0.3\textwidth}
    \includegraphics[width=\textwidth]{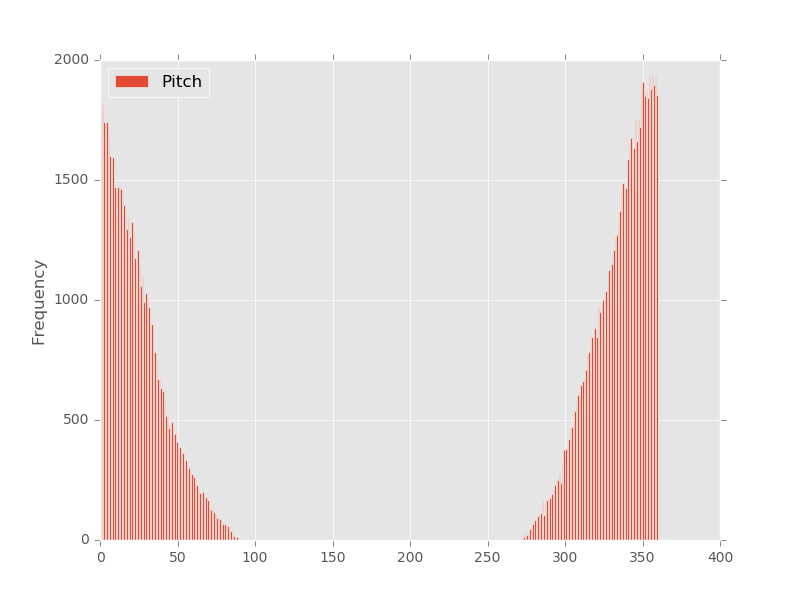}
  \end{minipage}
  \begin{minipage}[b]{0.3\textwidth}
    \includegraphics[width=\textwidth]{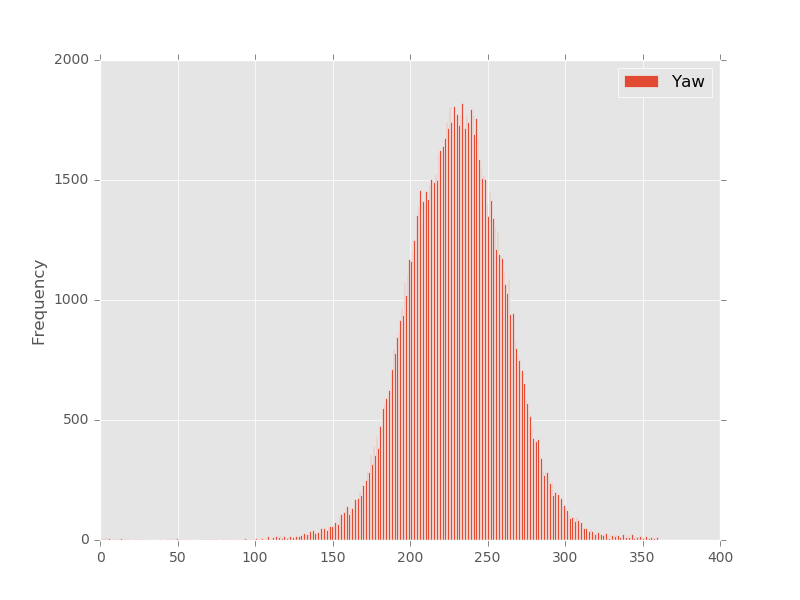}
  \end{minipage} \par
  \captionof{figure}{\textit{Left to right}: Histograms of roll, pitch and yaw in degrees.}
  \label{fig:orientation-hist}
\end{center}

Groundtruth pose distribution in our dataset aligns with the space of natural
human arm movement. We instructed all participants to use their right hands to
operate the controller, therefore there are more samples in the right half of
the image. We could easily flip the image vertically and use them as samples for
building left hand models.

\section{Additional Fields Multibox Detector}
\label{sec:model}

We define an extensible model architecture based on Single Shot Multibox Detector (SSD)
\cite{liu2016ssd} for egocentric hand detection and 6-DoF keypoint tracking
experiments. Our model \textbf{SSD-AF} supports adding arbitrary number of
\textbf{A}dditional \textbf{F}ields to
the output of each box prediction. Such additional fields can be used to
encode both regression and classification targets, for the cases of pose
regression models and binning models respectively.

\begin{center}
   \centering
   \includegraphics[width=0.95\textwidth]{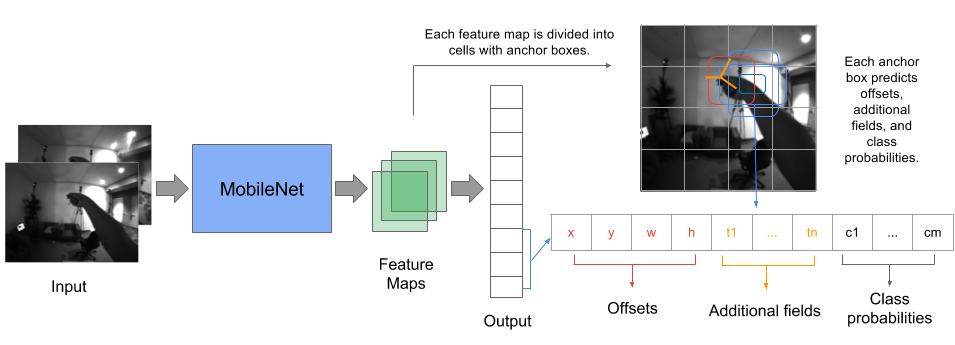}
   \captionof{figure}{MobileNet-SSD-AF architecture: we use MobileNet as
   the feature extractor network, and SSD with additional field output for
   detection and classification.}
   \label{fig:model}
\end{center}

We choose an object detection approach based on SSD for our task because it has
been shown to achieve good speed and accuracy tradeoff \cite{speedaccuracy}.
Additionally, SSD and similar approaches like YOLO \cite{redmon2016you} perform
detection and classification simultaneously and thus tend to be more
computationally efficient compared to two stage object detectors like Faster
RCNN \cite{girshick2015fast}. The SSD architecture also uses multi-scale feature
maps from the feature extractor network to enable detection at different scales.

The classic form of the SSD architecture is a
fully convolutional neural network that produces a fixed size collection of
bounding boxes and class scores for the presence of an object within the box.
This is followed by a non-maximum suppression step that chooses one or more
boxes with the highest class scores.The output collection of boxes are
represented as offsets to a set of heuristically chosen default boxes (similar
to anchor boxes in \cite{girshick2015fast}) of varying size and aspect ratio.
For each default box we have:
\begin{equation}
output = [ x_i, y_i, w_i, h_i, c_{i1}, \dots, c_{im} ] \text{ for default box } i \in \{1,\dotsc,n\}
\end{equation}
where $x_i$, $y_i$, $w_i$, $h_i$ represents the offsets, $c_{ij}$ represents
class probability for class $j$, and $n$ is the total number of matched
boxes.

During training, target offset and class probabilities are assigned to default
boxes whose overlap with groundtruth bounding boxes is above a given threshold.
The loss calculated between the output and target vectors has two part:
localization loss for the bounding box offsets, and classification loss for
associated class confidences. The total loss is the weighted sum of the
losses for the matched boxes and given by,
\begin{equation}
\begin{split}
  Loss = \frac{1}{n} [& L_{loc}(\langle x, y, w, h \rangle^{pred}_i,
                    \langle x, y, w, h \rangle^{gt}_i, x_i) +  \\
         &           \alpha L_{conf}(c^{pred}_{ij}, c^{gt}_{ij}, x_i)] 
         \quad \forall i \in \{1,\dotsc,n\}, j \in \{1,\dotsc,m\}
\end{split}
\end{equation}
$x_i$ is an indicator variable representing whether the $i$-th default
box was matched to a ground truth box.

In order to predict 6-DoF pose along with the bounding boxes, we make the
following modifications to the SSD architecture. We attach selected additional
fields to each of the default boxes such that each box's output now has $4$
offsets ($x, y, w, h$), $k$ additional fields ($t_1, t_2, \dots, t_k$) and $m$
class confidences ($c_1, c_2, \dots, c_m$). For example, if we use the
additional fields to predict the 2D keypoint corresponding to the controller tip
then $k=2$, where as if we predict the full 6-DoF pose with position in $xyz$
space and orientation in quaternions then $k=7$.  By default, we use $m=1$ for
all cases since we have only one object class corresponding to the hand holding
the controller. The total loss in SSD-AF is,
\begin{equation}
\begin{split}
  Loss = \frac{1}{n} [&L_{loc}(\langle x, y, w, h \rangle^{pred}_i,
                    \langle x, y, w, h \rangle^{gt}_i, x_i) +  \\
         &           \alpha L_{conf}(c^{pred}_{ij}, c^{gt}_{ij}, x_i) +
                     \beta L_{fields}(t^{pred}_{il}, t^{gt}_{il}, x_i)]  \\
         & \forall i \in \{1,\dotsc,n\}, j \in \{1,\dotsc,m\}, l \in \{1,\dotsc,k\}
\end{split}
\end{equation}

Note that additional fields can vary depending on whether the model predicts
regression targets such as 2D keypoint, 3D keypoint or full 6-DoF pose of the
controller tip, or classification targets such as discretized bins of angular
rotation. Additional field loss $L_{fields}$ is set according to the type of
target.
Additionally, each additional field can be encoded with respect
to the default box
coordinates, just as in case of the bounding box coordinates which are encoded
as offsets with respect to the default box. 

We use MobileNet \cite{howard2017mobilenets} as the feature extractor network
for SSD-AF. Our final model architecture is shown in Figure \ref{fig:model}.

\section{Experiments}
\label{sec:experiments}
\subsection{Experiment Setup}

We split the HMD Controller dataset into a training and
evaluation set based on users. In total we use 508,690 samples for training and
38,756 samples for testing. All metrics below are reported on the testing set.

Our models are implemented using Python and Tensorflow. The input images are
downsized from their original resolution of $640\times480$ to $320\times240$.
The images are preprocessed by normalization to a $[0, 1]$ range, and random
contrast and brightness perturbation is applied during training. We use a
MobileNet with depth multipler $0.25$ as our feature extractor. The ground truth
target vector is generated by assigning anchors that have greater than $50\%$
IOU with the groundtruth boxes. We use Smooth L1 loss \cite{girshick2015fast}
for localization and additional fields and binary cross entropy for
classification. We set the loss weights $\alpha$ and $\beta$ to be
$1.0$ in all our experiments.

The network is trained using stochastic gradient descent with ADAM optimizer
\cite{kingma2014adam}. As a post-processing
step, we perform non-maximum suppression on the output boxes to pick the box
with the highest class probability score. The final output consists of the
coordinates of the output box with the corresponding additional fields and class
confidences.

\subsection{Metrics}

We derive our metrics based on those defined in \cite{hodavn2016evaluation}. For a
sample in our testing set, we denote the
groundtruth 2D bounding box with $B^{gt}$, and a candidate 2D box as $B^{pred}$. We denote
class probability of the object in the candidate box being users'
right hand holding the controller as $c^{pred}$. Conversely, the probability of
candidate box being in background class is $1 - c^{pred}$.

\vspace{\baselineskip}
\noindent
\textbf{Detection Metrics}: We use mean average precision (mAP) as our main
metric for detection. The following algorithm is used to determine whether $B^{pred}$ is a true positive $TP$, false
positive $FP$, true negative $TN$, or false negative $FN$:
\[
  B^{pred} \text{ is }
\begin{cases}
    TN, & \text{if } B^{gt} \text{ does not exist and } c^{pred} < t_c\\
    TP, & \text{if } B^{gt} \text{ exists and } c^{pred} > t_c  \text{ and }
    IOU_{(B^{GT}, B^{pred})} > t_{IOU}          \\
    FN, & \text{if } B^{gt} \text{ exists and } c^{pred} < t_c \\
    FP, & \text{otherwise}\\
\end{cases}
\]
where $t_c$ is a selected threshold on class probability, and $t_{IOU}$ is a
selected threshold on the value of intersection over union (IOU) between the
groundtruth box and candidate box. In our results below, we
set $t_c$ to be $0.0001$. Unless stated otherwise, we set $t_{IOU}$ to be
$0.05$, which maps to a maximum of $92.4$ millimeter in position error. Finally precision is given by $tp / (tp + fp)$.

For models which predicts orientation in discrete bins, we also evaluate the mAP
of bin classification. mAP in bin classification is only computed on the true
positives.

\vspace{\baselineskip}
\noindent
\textbf{Pose Metrics}: For regression targets, we calculate the
mean average error (MAE) and root mean-squared error (RMSE) between the
groundtruth and predicted values. We report keypoint errors in image space
coordinate $u$, $v$ in pixels, and in camera space coordinate $x$, $y$, $z$ in
meters. For experiments which has orientation as a regression target, we report
orientation errors in camera space in degrees.

\subsection{3D Position Estimation}
First we present results on 3D pose estimation with SSD-AF. Our best model
for this task \textbf{SSD-AF-Stereo3D} uses a stacked stereo image pair as input to
the network and predicts boxes with 6 additional fields representing the 3D
position of the controller tip in both cameras ($t_u^1, t_v^1, t_z^1, t_u^2,
t_v^2, t_z^2$). Let the 3D position of the object keypoint be $P_{Cam}^o =
(P_x^o, P_y^o, P_z^o)$, and the projected keypoint be $u_o, v_o$ in image space.
We encode the offset of $u_o, v_o$ with respect to the anchor box as $t_u$ and
$t_v$, and $P_z^o$ with respect to the box height as $t_z$.  We have:
\begin{equation}
  t_u  = (u_o - x_a) / w_a, \quad
  t_v  = (v_o - y_a) / h_a, \quad
  t_z  = P_z^o / h_a
\end{equation}
where $x_a, y_a$ represent the default box $a$'s center and $w_a, h_a$ represent
its width and height.

We show the qualitative results of our SSD-AF-Stereo3D model in Figure
\ref{fig:3dof-results}.

\begin{center}
  \centering
  \begin{minipage}[b]{0.49\textwidth}
    \includegraphics[width=\textwidth]{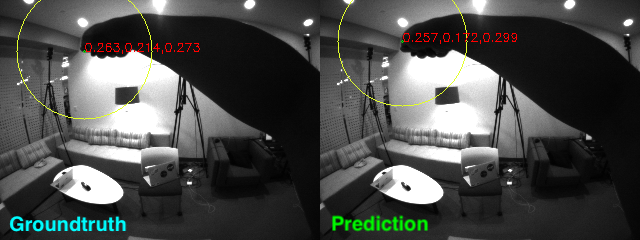}
  \end{minipage}
  \begin{minipage}[b]{0.49\textwidth}
    \includegraphics[width=\textwidth]{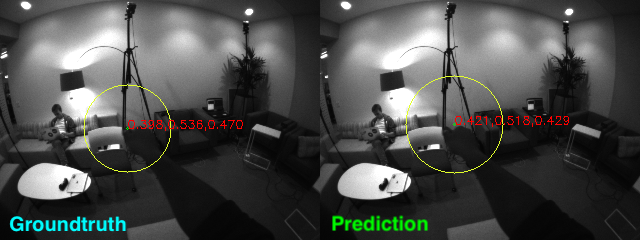}
  \end{minipage}
  \par 
  \begin{minipage}[b]{0.49\textwidth}
    \includegraphics[width=\textwidth]{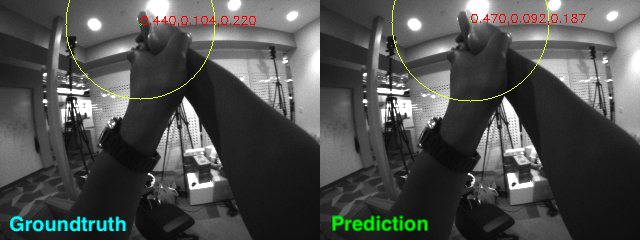}
  \end{minipage}
  \begin{minipage}[b]{0.49\textwidth}
    \includegraphics[width=\textwidth]{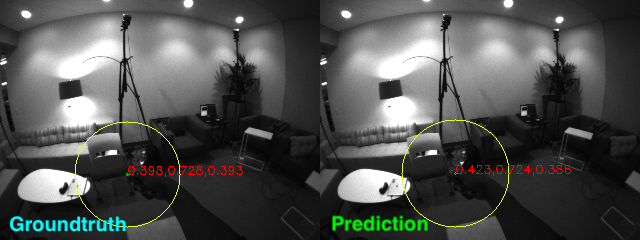}
  \end{minipage}
  \par
  \captionof{figure}{Prediction results of SSD-AF-Stereo3D on sample test
  set images. Groundtruth 3D position is visualized on the left half of each
  image, and predicted position is visualized on the right. The size of the
  overlay circle is inversely proportional to $P_z^o$. Our model performs within
  centimeter accuracy despite extremely challenging lighting conditions and
  complex situations such as user holding the
  controller with both hands.}
  \label{fig:3dof-results}
\end{center}

Quantatively, we compare our model with two other baseline model:
\begin{itemize}
  \item \textbf{SSD-AF-2D}: Model input is one of the stereo images. Additional
  fields output are $(t_u, t_v)$.
\item \textbf{SSD-AF-3D}: Model input is one of the stereo images. Additional
  fields output are $(t_u, t_v, t_z)$.
\end{itemize}

The evaluation results for bounding box and class prediction are shown in table
\ref{tab:results_bbox}, and the results for 2D, 3D and 6-DoF tracking are shown
in table \ref{tab:results_tracking}. It can be seen that the precision of the
models are generally high, indicating good bounding box prediction performance.
SSD-AF-Stereo3D achieved the best bounding box precision of $0.9534$. It
also achieves the lowest UV prediciton MAE of $8.45$ pixels and 3D keypoint
prediction MAE of $33.5$ millimeters. Additionally, this model can run on one big mobile CPU
at 30fps.

\begin{center}
\centering
  \begin{tabular}{|l|c|c|c|}
\hline
  Model & mAP @0.05IOU & mAP @0.25IOU & mAP @0.5IOU \\ \hline
  SSD-AF-2D      &  0.9140         & 0.8469 & 0.5180       \\ \hline
  SSD-AF-3D      &  0.9380         & 0.8761 & 0.5131      \\ \hline
  SSD-AF-Stereo3D      & \bfseries 0.9534          & \bfseries 0.9533 & \bfseries 0.7767       \\ \hline
\end{tabular}
\captionof{table}{Detection mAP of 3D position models. }
\label{tab:results_bbox}
\end{center}

\begin{center}
\centering
\begin{tabular}{|l|c|c|c|c|c|c|}
\hline
\multirow{2}{*}{Model} & \multicolumn{2}{c|}{Position ($uv$)} &
  \multicolumn{2}{c|}{Position ($xyz$)} & \multicolumn{2}{c|}{Latency (ms)} \\ \cline{2-7}
 & MAE & RMSE & MAE & RMSE & Mobile & Titan X \\ \hline
SSD-AF-2D & 12.41 & 30.01 & - & - & 30.140 & 6.378 \\ \hline
SSD-AF-3D & 10.23 & 24.38 & 0.0493 & 0.0937 & 30.649 & 6.303 \\ \hline
SSD-AF-Stereo3D & \bfseries 8.45 & \bfseries 23.25 & \bfseries 0.0335 &
  \bfseries 0.0776 & 31.768 & 6.512 \\ \hline
\end{tabular}
\captionof{table}{Pose predition errors and latency: Position errors is measured
  in pixels in $uv$ space, and in meters in $xyz$ space. Mobile latency is measured in
  milliseconds on
  a Pixel 2 mobile phone using only 1 big CPU core, and on a desktop Titan X
  GPU. Note that we can run SSD-AF-2D on both images in the stereo pair, and
  triangulate with camera extrinsics to compute the 3D pose. This would
  effectively double the runtime.}
\label{tab:results_tracking}
\end{center}

The higher $uv$ prediction performance of the
SSD-AF-3D model compared to the SSD-AF-2D model indicates that
adding $t_z$ to the target helps bounding box and 2D keypoint
prediction as well. Our observation is aligned with the theory that
adding additional supervised information helps neural networks learn.

SSD-AF-Stereo3D model performs the best indicates that the model is able
to infer positional information better using both stereo images as input.
Interestingly, we also observed that models that use stereo input but only
predict $(t_u, t_v, t_z)$ in one of the images (instead of both) do not
out-perform single image models such as SSD-AF-3D.

\subsection{Orientation and 6-DoF Prediction}
Second we also present results on orientation and 6-DoF pose estimation with
SSD-AF. Recent notable work on 6-DoF pose estimation typically uses one of
two methods: regression or discrete binning. Regression models such as in
\cite{mahendran20173d} predicts object poses directly. Orientation can be represented in
either Euler angles or quaternions for regression. Discrete binning model such
as in \cite{poirson2016fast} splits the possible 6-DoF space into a number of discrete
bins, or \textit{Views} as in \cite{kehl2017ssd}. Pose estimation then becomes a
classification problem of assigning the correct viewspace bin to the sample. 

We implemented both approaches with SSD-AF model:
\begin{itemize}
  \item \textbf{SSD-AF-Stereo6D-Quat}: This model takes a stacked stereo pair as input and
    predict boxes with 14 additional fields that
    represent the full 6 DoF pose ($t_u, t_v, t_z, qx, qy, qz, qw$) of the controller
    in both images. $qx, qy, qz, qw$ is the quaternion representation of
    orientation. 
  \item \textbf{SSD-AF-Stereo6D-Euler}: This model is similar to the one
    above besides that orientation is represented by 3 values $\alpha, \beta,
    \gamma$ in pitch, yaw, roll direction in Euler angle.
  \item \textbf{SSD-AF-Binned}: Instead of regression target, this model
    outputs $b$ additional fields in class probabilities ($tc_1, \dotsc,
    tc_b$). $tc_i$ corresponds to the $i$-th orientation bin. In our
    experiments, we split the full orientation space equally into bins.
  \item \textbf{SSD-AF-3D-Binned}: Similar to above but also predicts
    ($t_u, t_v, t_z$) in addition to orientation bins.
  \item \textbf{SSD-AF-3D-AxisBinned}: Similar to above but orientation is
    binned per axis.
\end{itemize}

Additionally, we also test the \textbf{SSD-AF-MultiplePoint} model,
which outputs additional fields ($t_u, t_v, t_z$) for $4$ keypoints for each
image in the stereo pair, yielding a total of $24$ additional fields. The
additional keypoints are chosen to correspond to other keypoints on the
controller which are not co-planar. We compute orientation from the $4$
keypoints by fitting a plane to the predicted keypoints and computing the
orientation of the plane in camera space.

The results of these experiments are shown in Table
\ref{tab:results_orientation} and Table \ref{tab:binning_map}. 

\begin{center}
\centering
\begin{tabular}{|l|C|C|C|c|}
\hline
  \multirow{2}{*}{Model} & \multicolumn{3}{c|}{Orientation MAE} &
  \multicolumn{1}{c|}{Position MAE} \\ \cline{2-5}
  & Yaw & Pitch & Roll & $xyz$ \\ \hline
  SSD-AF-Stereo3D & - & - & - & \bfseries 0.0335 \\ \hline
  SSD-AF-Stereo6D-Quat & 0.3666 & 1.4790 & 0.6653 & 0.0521 \\ \hline
  SSD-AF-Stereo6D-Euler & 0.3630 & 1.5840 & 0.7334 & 0.0448 \\ \hline
  SSD-AF-MultiplePoint & 0.3711 & 1.108 & 1.203 & 0.0452\\ \hline
  SSD-AF-3D-AxisBinned (20$\times$3 bins) & \bfseries 0.1231 & \bfseries 0.8594 & \bfseries 0.5256 &
  0.0503 \\ \hline
\end{tabular}
  \captionof{table}{MAE of orientation prediction models. Errors are measured in
  radians.}
  \label{tab:results_orientation}
\end{center}

\begin{center}
\centering
  \begin{tabular}{|l|c|c|}
\hline
    Model  & \makecell{Orientation Bins \\ mAP} & \makecell{Position \\MAE} \\ \hline
    SSD-AF-Binned (27 bins) & 0.6538 &     - \\ \hline
    SSD-AF-Binned (512 bins) & 0.3627   &  -\\ \hline
    SSD-AF-3D-Binned (27 bins) & 0.6412 &  0.04760 \\ \hline
    SSD-AF-3D-Binned (512 bins) & 0.3801 & 0.07167  \\ \hline
    SSD-AF-3D-AxisBinned-Yaw (20 bins) & 0.4480 &   0.05124 \\ \hline
    SSD-AF-3D-AxisBinned-Pitch (20 bins) & 0.3592 &   0.04975 \\ \hline
    SSD-AF-3D-AxisBinned-Roll (20 bins) & 0.5532 & 0.04413 \\ \hline
\end{tabular}
\captionof{table}{Classification mAP of bining models. }
\label{tab:binning_map}
\end{center}

SSD-AF-3D-AxisBinned performs the
best across the board with the lowest numbers in all three directions. Note that
this model predicts orientation around only one of these directions at a time
instead of simultaneously as in the case of the others. In general, binning
models models perform better than regression models on orientation. Our models
with 512 bins achieves binning precision $38\%$, which is much higher than
chance.

Among regression models, the quaternion encoding of
SSD-AF-Stereo6D-Quat performs slightly better than the Euler angle encoding of
SSD-AF-Stereo6D-Euler. This observation is different from results in
\cite{mahendran20173d}. SSD-AF-MultiplePoint outperforms SSD-AF-Stereo6D models in the
pitch direction, but fails short in the roll direction.

Finally, all 6-DoF models perform slightly worse on 3D position prediction
compared to SSD-AF-Stereo3D. We conjecture that this is due to our mobile
friendly models running out of capacity for predicting both the position and
rotation.

\section{Conclusion and Future Work}

We have presented approaches for efficient 6-DoF tracking of handheld
controllers on mobile VR/AR headsets. Our methods use stereo cameras on the
headset, and IMU on 3-DoF controllers as input.  The HMD Controller dataset
collected for this work consists of over 540,000 stereo pairs of fisheye images
with markerless 6-DoF annotation of the controller pose. The 6-DoF annotation is
automatically collected with a Vicon motion capture system and
has sub-millimeter accuracy. Our dataset covers a diverse
user base and challenging environments. To the best of our knowledge this is the
largest dataset of its kind.

We have demonstrated that our SSD-AF-Stereo3D model achieves a low positional
error of 33.5mm in 3D keypoint tracking on our dataset. It can run on a single
mobile CPU core at 30 frames per second. We have also presented results on
end-to-end 6-DoF pose prediction under strict computational constraints.

Our future work includes improving orientation prediction results. We believe
our models can be further improved by encoding orientation to be invariant to
default box locations. Objects with the same orientation may have different
appearance in different parts of the image due to camera projection. Instead of
asking the network to learn the projection, we can explore using
projection-adjusted orientation as groundtruth, such that objects with the same
apperanace always correspond to the same orientation label.

Another interesting research direction is to apply temporal and contextual
information to our models. Currently all our models predict object poses on a
frame-by-frame basis. Adding temporal filtering or using a RNN could
significantly speed up tracking. Motion priors for different types of
interaction can also be added to further improve tracking quality.

\clearpage

\bibliographystyle{splncs}
\bibliography{egbib}
\end{document}